\documentclass[letterpaper]{article} 
\usepackage{aaai2026}  
\usepackage{times}  
\usepackage{helvet}  
\usepackage{courier}  
\usepackage[hyphens]{url}  
\usepackage{graphicx} 
\usepackage{natbib}  
\usepackage{caption} 
\usepackage{algorithm}
\usepackage{algorithmic}
\usepackage[most]{tcolorbox}
\usepackage{xcolor}
\usepackage{listings}
\usepackage{amsmath, amssymb}

\usepackage{multirow}
\usepackage{makecell}
\usepackage{booktabs}

\usepackage{newfloat}
\usepackage{listings}
\DeclareCaptionStyle{ruled}{labelfont=normalfont,labelsep=colon,strut=off} 
\floatstyle{ruled}
\newfloat{listing}{tb}{lst}{}
\floatname{listing}{Listing}
\newcounter{corrfn}
\makeatletter  
\def\corresponding{%
  \ifnum\value{corrfn}=0%
    \footnote{They are co-corresponding authors}%
    \setcounter{corrfn}{\value{footnote}}
  \else
    \footnotemark[\value{corrfn}]%
  \fi%
}
\makeatother
\title{Ground What You See: Hallucination-Resistant MLLMs via Caption Feedback, Diversity-Aware Sampling, and Conflict Regularization}
\author{
    Miao Pan\textsuperscript{\rm 1}\equalcontrib, 
    Wangjie Gan\textsuperscript{\rm 1}\equalcontrib, 
    Jintao Chen\textsuperscript{\rm 1}\corresponding,
    Wenqi Zhang\textsuperscript{\rm 1}, \\
    Sun Bing\textsuperscript{\rm 2},
    Jianwei Yin\textsuperscript{\rm 1},
    Xuhong Zhang\textsuperscript{\rm 1,3}\corresponding
}
\affiliations{
    \textsuperscript{\rm 1}School of Software Technology, Zhejiang University\\
    \textsuperscript{\rm 2}National Certification Technology (Hangzhou) Co., Ltd\\
    \textsuperscript{\rm 3}Ningbo Global Innovation Center, Zhejiang University\\
    \{chenjintao,zhangxuhong\}@zju.edu.cn

}

\usepackage{bibentry}

\begin{document}

\maketitle

\begin{abstract}
Multimodal large language models (MLLMs) have achieved significant results in various tasks, but their practical application is still severely constrained by hallucination issues, which are particularly prominent in reinforcement learning (RL) optimization processes. This paper systematically analyzes the causes of hallucinations in MLLM under RL training, identifying three key factors: (1) The model relies heavily on chained visual reasoning to guide decision-making during RL training. Thus, error and irrelevant information in visual reasoning can easily cause hallucinations, including inaccurate initial visual descriptions that anchor subsequent inferences to incorrect information, as well as redundant and broad inferential information; (2) Insufficient exploration diversity during the policy optimization phase, causing the model to output overly confident results; (3) The destructive conflict between different samples during optimization is a key factor that leads to false associations and unstable parameter updates. To address these issues, we propose a solution framework comprising three core modules. First, to improve the accuracy of visual localization, we add planning and caption stages before thinking and answer stages. To enhance initial visual descriptions ability, we allow LLMs to respond based solely on the caption and provide corresponding caption reward based on the quality of the response. Second, to enhance exploration capabilities, we classify samples based on the mean and variance of the reward distribution and select samples with high reward variance for training, thereby increasing the model's focus on diverse samples. Finally, to mitigate conflicts between training samples, we identify neural tangent kernel (NTK) similarity as the key factor. Rather than minimizing it uniformly, we regulate NTK similarity by grouping sample pairs based on a similarity threshold. An InfoNCE loss is then applied to pull dissimilar pairs closer and push overly similar ones apart, guiding interactions toward a balanced range. The experimental results demonstrate that the proposed method significantly reduces the hallucination rate and effectively improves the inference accuracy of MLLMs.
\end{abstract}

\begin{links}
    \link{Code}{https://github.com/ZJU-OmniAI/OMNEX-VL}
\end{links}

\section{Introduction}

Multimodal large language models (MLLMs)~\cite{hurst2024gpt,team2024gemini} have demonstrated strong performance across a range of multimodal tasks. However, despite their impressive performance, these models are prone to hallucination~\cite{dona2025bettercheck}, producing responses that are linguistically fluent but factually inconsistent with the visual input~\cite{leng2024mitigating}. This limitation poses a significant challenge for real-world deployment, particularly in safety-critical applications.

While hallucination is a common issue across MLLMs, such problems can become even more pronounced during reinforcement learning (RL) optimization~\cite{Kaelbling_Littman_Moore_2018}. In particular, RL-based training tends to amplify \textit{Flaw Repetition}, where models fall into loops of semantically redundant reasoning, and introduce \textit{Think-Answer Mismatch}, where final answers deviate from preceding thought processes~\cite{yao2025reasoning}. To better understand the underlying factors behind these behaviors, we identify three major challenges that arise in RL training.

\begin{figure*}[h]  

  \centering

  \begin{minipage}{\linewidth}

    \includegraphics[width=\linewidth]{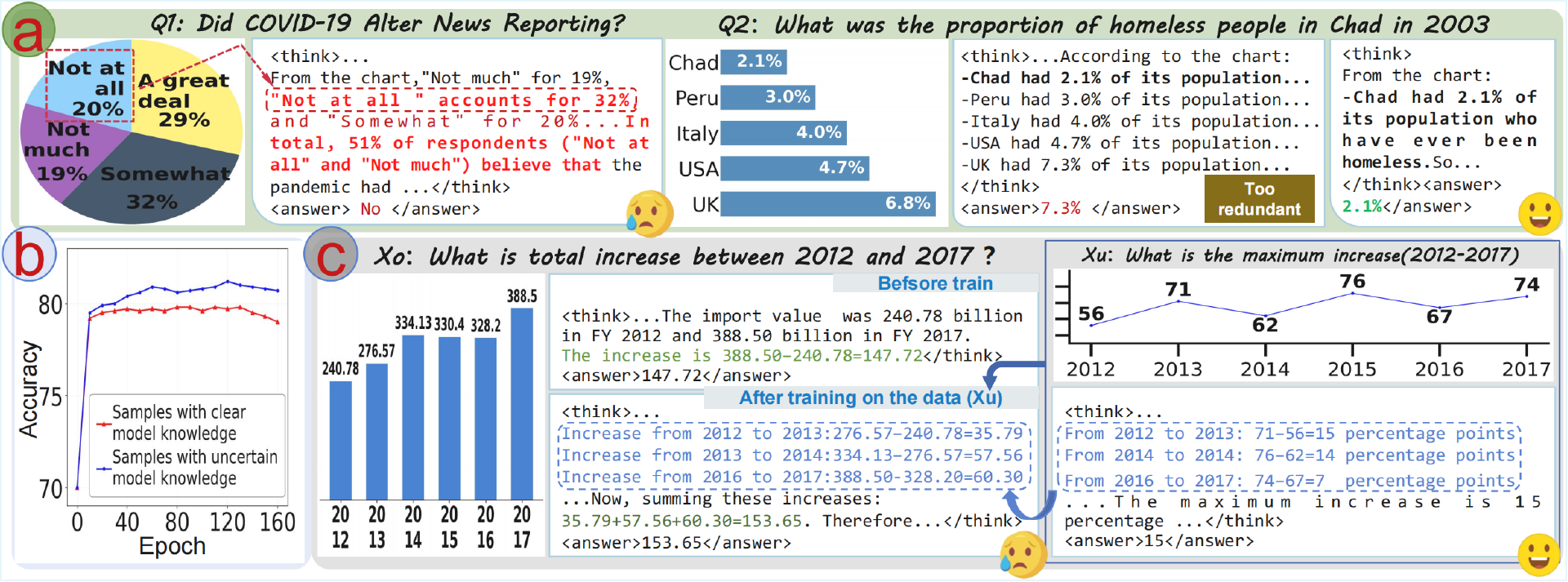}

    \caption{Three Types of Hallucination in RL-Tuned Multimodal LLMs: Visual Misinterpretation, Limited Exploration, and Sample Conflict}

    \label{fig:methodfig1}

  \end{minipage}

\end{figure*}

First, in multimodal domains, reasoning steps are essential as they guide the interpretation of visual information~\cite{yao2025reasoning}. RL training may introduce hallucinations due to \textbf{Visual Misinterpretation}~\cite{christiano2017deep}. Hallucination occurs when the initial visual description inaccurately reflects the visual content, leading the model to propagate this misinformation through subsequent reasoning and ultimately generate incorrect answers (Figure~\ref{fig:methodfig1}(a)-Q1). Another form arises when the model broadly or imprecisely attends to the input, leading it to produce redundant or irrelevant reasoning that is mistakenly treated as key evidence. This misattribution obscures truly informative cues and ultimately results in incorrect conclusions ( Figure~\ref{fig:methodfig1}(a)-Q2).

Second, during RL training, \textbf{Limited Exploration} diversity (MLLM output distribution) may lead to hallucinations. As shown in Figure~\ref{fig:methodfig1}(b), samples with higher reward variance consistently achieve better generalization performance compared to those with lower reward variance. This suggests that focusing primarily on samples with lower reward variance tends to reinforce dominant output patterns and reduce output diversity. As a result, the model's exploration space becomes narrower, increasing the risk of hallucinations caused by overfitting and the over-inference leading to false associations, especially in ambiguous contexts.

Third, training \textbf{Samples Conflict} may contribute to hallucinations in MLLMs. As shown in Figure~\ref{fig:methodfig1}(c), updating the model on one sample $x_u$ may unintentionally affect the model’s predictions on other, unrelated sample $x_o$. Such conflicts may introduce spurious associations between unrelated samples in the model's reasoning process, leading to overconfident yet incorrect predictions for inputs. This significantly heightens the risk of hallucinations across diverse scenarios. However, the key factors underlying such conflicts remain underexplored in existing research.

To address these hallucination challenges, we introduce three key components. First, to ensure that model reasoning remains grounded in visual evidence, we add \textbf{planning and caption stages} before thinking and answer stages. In addition, we introduce a \textbf{Caption Reward}, which quantitatively measures the consistency between the generated caption and the visual input. This is achieved by verifying whether a language model can correctly answer the question using only the generated caption and the question itself. Secondly, to mitigate false correlations caused by insufficient exploration in RL, we analyzed how reward signals influence the model's output distribution. We found that \textit{positive rewards make the distribution more peaked, while negative rewards make the distribution flatter}. Based on this, we determined that medium samples (high reward variance) are the most valuable for training, as they enable effective exploration in RL, guiding the model from uncertainty toward confident predictions. To enhance learning diversity, we retained only these \textbf{high-variance samples} for RL updates. Finally, to mitigate sample conflicts , we identify \textbf{NTK similarity}~\cite{jacot2018neural} as the primary factor. When NTK similarity is too high, conflicting samples can exert excessive influence on each other; when too low, the influence among mutually beneficial samples becomes too weak. Therefore, instead of uniformly minimizing NTK similarity, we propose to regulate it by grouping sample pairs based on a threshold: pairs with excessively high similarity are treated as negatives, while those with low similarity are treated as positives. An \textbf{InfoNCE loss} is then applied to pull positive pairs closer and push negative pairs apart, guiding NTK similarity into a balanced range.

\begin{itemize}
    \item We introduce a Visual-Grounded Reasoning mechanism by adding planning and caption stages before the thinking and answer stages. In addition, we incorporate a Caption Reward that quantifies visual-text alignment by verifying whether the caption alone enables a language model to answer the question correctly.

    \item We address hallucinations caused by limited exploration diversity in RL training by categorizing samples according to the mean and variance of their reward scores and selecting those with high reward variance for training.
    
    \item We introduce a Conflict-Aware Regularization mechanism by measuring the NTK similarity between samples. In addition, we incorporate an InfoNCE loss that guides the similarity values—determined by a threshold $\tau$—toward an appropriate range.
\end{itemize}

\section{Related work}
\paragraph{Reinforcement Learning for MLLMs.}
Reinforcement learning is rapidly advancing multimodal large models, enhancing their reasoning and task performance. Researchers, inspired by the format and accuracy reward mechanisms in DeepSeek R1-Zero's Equation 13, are now trying to apply this successful approach to multimodal tasks~\cite{pan2025medvlm,zhou2025r1,meng2025mm,liu2025othink}. Recent applications demonstrate diverse strategies for integrating RL, particularly for complex, multi-step problems and multimodal information. For instance, Vision-R1~\cite{huang2025vision} used Progressive Thinking Suppression Training (PTST) to extend Chain-of-Thought (CoT) and decouple rewards, while R1-VL~\cite{zhang2025r1} introduced StepGRPO with StepRAR and StepRVR for consistency in complex tasks. For multimodal integration, LMM-R1~\cite{peng2025lmm} used a two-stage training (text then image-text), and R1-Onevision~\cite{yang2025r1} formalized image descriptions to integrate visual information. In terms of cross-modal reward design, Video-R1~\cite{feng2025video} added a temporal consistency reward for video understanding. Specialized applications include Reason-RFT, which customizes rewards for mathematical problems, Q-Insight~\cite{li2025q} integrated verifiable score accuracy for image quality assessment and additional~\cite{lu2025ui,liu2025seg,pan2025metaspatial,kang2025gflowvlm,deng2025boosting}. However, previous studies have largely ignored the illusions caused by visual extraction errors and how to design reward signals to guide the model to generate accurate and vision-based captions. This limits the effectiveness of reinforcement learning-based alignment in multimodal settings.
\paragraph{Hallucination in MLLMs.}
Hallucination, generating fluent but factually incorrect content, remains a critical challenge in large-scale models, particularly in multimodal and multi-step reasoning. In vision-language models, hallucination is amplified as attention to visual tokens degrades~\cite{liu2025more}, revealing a reliance on language priors. This aligns with findings that high-reward samples disproportionately influence updates, while semantically overlapping, moderately incorrect answers propagate misleading gradients~\cite{zhu2025surprising}. Existing mitigation strategies include visual grounding via retrieval~\cite{lee2025retrieval}, hallucination-aware decoding~\cite{leng2024mitigating}, and representation-space filtering~\cite{ghosh2024visual}. Crucially, GRPO exacerbates hallucination in MLLMs by reinforcing semantically plausible but visually ungrounded responses, especially with sparse or ambiguous supervision.

\section{Preliminary}
\textbf{Group Relative Policy Optimization (GRPO)}~\cite{shao2024deepseekmath}. We apply GRPO to optimize the MLLM using fine-grained rewards. In each training iteration, for a sample $q$, the model $\pi_\theta$ generates $G$ candidate responses, where each $y_i = (y_{i,1}, \dots, y_{i,T})$ is a sequence of tokens. For each candidate, a scalar reward $r_i$ is assigned based on answer quality. The sequence-level reward is normalized to obtain a per-token advantage $\text{A}_{i,t} = \frac{r_i - \text{mean}(r)}{\text{std}(r)}$. The likelihood ratio $\rho_{i,t} = \frac{\pi_\theta(y_{i,t} | q, y_{i,<t})}{\pi_{\text{ref}}(y_{i,t} | q, y_{i,<t})}$ is used to compare the current and reference policies at each token. The loss is as follows:
\begin{equation}\label{loss}
\begin{aligned}
&\mathcal{L}_{\text{GRPO}}(\theta) = \mathbb{E}_{q, \{y_i\}} \sum_{i=1}^{G} \frac{1}{|y_i|} \sum_{t=1}^{|y_i|} [\beta\, \mathbb{D}_{\text{KL}}(\pi_{\theta} \| \pi_{\text{ref}})
\\
& + \min \left( \rho_{i,t} \cdot \text{A}_{i,t},\ \text{clip}(\rho_{i,t}, 1 - \varepsilon,\ 1 + \varepsilon) \cdot \text{A}_{i,t} \right)],
\end{aligned}
\end{equation}
where $\beta$ controls the KL penalty and $\varepsilon$ is the clipping threshold. This objective encourages the model to generate high-reward responses while constraining deviation from the reference policy.

\begin{figure*}[h]
    \centering
    \includegraphics[width=1\linewidth]{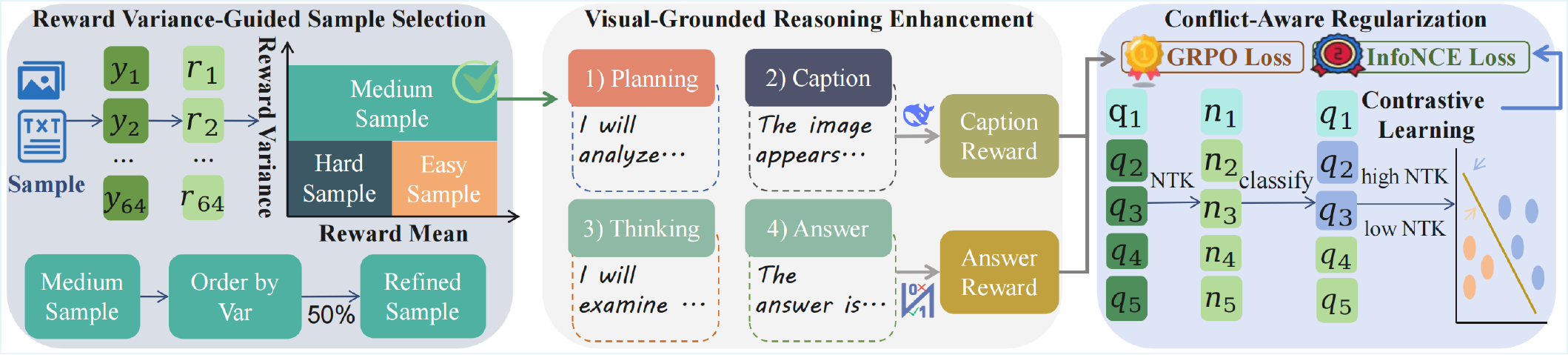}
    \caption{The proposed framework for robust visual reasoning, composed of three components: Reward Variance-Guided Sample Selection, Visual-Grounded Reasoning Enhancement, and Conflict-Aware Regularization.}
    \label{fig:overview}
\end{figure*}
\section{Method}
To address hallucination in RL training, we introduce three key components, illustrated in Figure~\ref{fig:overview}. First, to ensure reasoning remains grounded in visual evidence, we extend the standard reasoning process by introducing two additional stages—planning and caption—before the thinking and answer stages. We further propose a Caption Reward that evaluates the consistency between generated captions and visual input by testing whether an LLM, given the caption and question, can produce the correct answer. Second, we observe that in RL training, positive rewards sharpen the output distribution, while negative rewards flatten it. This reveals that medium samples—those with high reward variance—are key to effective exploration. They initially induce flattening when the model is uncertain, and later lead to sharpening as predictions improve. This transition from exploration to confident exploitation reflects an ideal learning trajectory. To select high-quality samples that support this process, we compute the reward variance across multiple responses per input and retain only the high-variance samples for RL. Finally, we observe that harmful gradient conflicts are closely tied to NTK similarity~\cite{jacot2018neural} between training samples. To mitigate such interference without suppressing beneficial interactions, we regulate rather than eliminate NTK similarity. For each sample, we compute its NTK similarity with others and partition pairs based on a threshold. An InfoNCE loss is then applied to encourage alignment among dissimilar pairs and discourage convergence among overly similar ones, guiding NTK similarity into a balanced range.

\subsection{Visual-Grounded Reasoning Enhancement}

\textbf{Redefine the Reasoning Paradigm}. As illustrated in Figure~\ref{fig:methodfig1}(a), a primary source of hallucinations arises when the model fails to accurately localize relevant visual cues. To strengthen visually grounded reasoning, we add planning and caption stages before the thinking and answer stages. In the planning stage, we perform early localization of problem-relevant visual regions. The caption stage generates concise textual descriptions for these regions, providing focused intermediate guidance. The thinking stage incorporates these descriptions into multi-step reasoning, while the answer stage synthesizes them into the final response.

\textbf{Caption Reward}. To address the issue of hallucinations caused by inconsistencies between the early-generated text signals and the visual signals, we propose the \textit{Caption Reward}. This mechanism ensures that the generated text signals are consistent with the visual signals, thereby preventing erroneous reasoning and mitigating hallucinations in later stages. First, we extract the textual caption generated by the model, which serves as the linguistic description of the visual input. This caption is then combined with the question and fed into a separate large language model (LLM) to generate a final answer. If the LLM is able to correctly answer the question only based on the caption, it implies that the model’s extracted caption is both accurate and effective in representing the visual content. If the LLM answers correctly using only the caption, a positive reward is given; otherwise, the model receives zero reward.

\subsection{Reward Variance-Guided Sample Selection}
To investigate how reinforcement learning drives output diversity in MLLMs, we examine how different training samples influence the model's distributional dynamics. Within a batch, samples naturally vary in how well the model initially understands them—reflected in their reward values. We categorize these samples into two types: \textbf{samples with clear model knowledge}, which receive high rewards, indicating that the model assigns high confidence to the correct answer; and \textbf{samples with uncertain model knowledge}, which receive low rewards, suggesting that the model incorrectly assigns high confidence to a wrong answer.

To quantify how the model’s output distribution changes for each sample, we generate an answer for every training instance, assign it a reward score based on output quality, and perform mean normalization to compute the per-sample advantage \( A_{i,t} \). By taking the gradient of GRPO loss with respect to the logits \( z_v \), we obtain:
\begin{equation}\label{g}
\frac{\partial \mathcal{L}_{i}}{\partial z_v}
\propto
\begin{cases}
-\pi_{y_t} \cdot (1 - \pi_{y_t}) \cdot A_{i,t} & \text{if } v = y_t \\
\pi_v \cdot \pi_{y_t} \cdot A_{i,t} & \text{if } v \ne y_t
\end{cases}
\end{equation}

where \( z_v \) denotes the logit of token \( v \), \( \pi_v \) is its corresponding softmax probability, \( y_t \) is the sampled token at position \( t \), and \( A_{i,t} \) is the normalized advantage of sample \( i \) at time \( t \). This expression reveals that the \textit{sign of \( A_{i,t} \)} governs the direction of change in the output distribution: when \( A_{i,t} > 0 \), corresponding to samples with clear model knowledge, the gradient update \textbf{sharpens the distribution}—it increases the probability of the sampled token while decreasing those of others. Conversely, when \( A_{i,t} < 0 \), corresponding to samples with uncertain model knowledge, the gradient update \textbf{flattens the distribution}—it lowers the probability of the sampled token while increasing those of unsampled tokens. 

If a batch contains mostly high-reward samples, repeated sharpening can lead to \textit{premature overfitting} and degraded generalization. Conversely, if dominated by low-reward samples, the model may enter a regime of \textit{persistent uncertainty}, yielding overly flat distributions and ineffective exploration. The most valuable training signals arise from \textbf{samples with high reward variance}—those the model sometimes gets right and sometimes wrong. Initially, such samples receive negative rewards that flatten the distribution due to incorrect high-confidence outputs. As training progresses and predictions improve, positive rewards gradually sharpen the distribution. This transition from exploration to confident exploitation reflects an ideal learning trajectory.

\textbf{Sample Classification}. For each input query, we generate 64 responses and compute the mean and variance of their computed reward scores. Based on these statistics, samples are categorized into three types: (i) easy samples with high mean and low variance, (ii) hard samples with low mean and low variance, and (iii) medium samples with high variance. To prioritize informative training signals, we use reward variance as a selection score and retain only the top 50\% high-variance samples for RL training.

\subsection{Conflict-Aware Regularization}

To understand how updates on one sample influence others, we investigate  how training samples \( (x_u, y_u) \) conflict with prior knowledge associated with \( x_o \) through the NTK-based learning dynamics. Specifically, we leverage formulation:
\begin{equation}\label{ntk}
\begin{aligned}
    &\Delta \log \pi^t(\mathbf{y}_u \mid \mathbf{x}_o) \\=& \eta\cdot \mathcal{A}^t(x_o)\cdot \mathcal{K}^t(x_o, x_u) \cdot \nabla_z \log \pi^t(y_u \mid x_u) \cdot A_{u,t} \\+& \mathcal{O}\left(\eta^2 \left\| \nabla_\theta \mathbf{z}(\mathbf{x}_u) \right\|^2_{\mathrm{op}} \right),
\end{aligned}
\end{equation}

where \( \eta \) is the learning rate, and \( \Delta \log \pi^t(\mathbf{y}_u \mid \mathbf{x}_o) \) denotes the change in log-probability at \( x_o \) after a policy update using sample \( (x_u, y_u) \). Here, \( \mathcal{A}^t(x_o) = I - \mathbf{1} \cdot \pi^t(x_o)^\top \) is the Jacobian of log-probability w.r.t. logits, and \( \mathcal{K}^t(x_o, x_u) = \nabla_\theta z(x_o)^\top \nabla_\theta z(x_u) \) is the empirical neural tangent kernel.  
The vector \( \nabla_z \log \pi^t(y_u \mid x_u) \) denotes the derivative of the log-probability with respect to the logits \( z \), and \( A_{u,t} \) is the advantage. The second-order term is typically small under gradient clipping.

The conflict effect is primarily governed by the NTK term $\mathcal{K}^t(x_o, x_u)$. When the NTK term is high, even for unrelated queries, reward-driven updates at $(x_u, y_u)$ can substantially alter the model's predictions at $x_o$. Consequently, the model may begin to exhibit overgeneralization, where responses meant for specific contexts are inappropriately applied elsewhere. Notably, the NTK similarity is quite different from true semantic similarity. Therefore, model may produce confident yet incorrect outputs that are similar via NTK but unrelated in meaning.

\textbf{Conflict Mitigation via InfoNCE Loss}. To mitigate such unintended conflict, we propose to regulate NTK similarity rather than eliminate it entirely. Excessively reducing NTK similarity between training samples may suppress beneficial interactions and hinder generalization. To strike a balance, we aim to regulate NTK-based similarity rather than minimize it uniformly. Since the NTK is difficult to compute directly, we approximate NTK similarity using the cosine similarity between the log-probability gradients of two samples at the final layer. For each sample, we compare it against other samples in the same batch and classify them into positive or negative pairs based on a threshold~$\tau$: samples with lower similarity are treated as positives (encouraged to \textbf{align}), while those with excessive similarity are treated as negatives (encouraged to \textbf{diverge}). The InfoNCE loss is then applied to adjust these relations, guiding the NTK similarity values into a balanced range that reduces harmful interference while preserving cooperative signals.
\begin{equation}
\mathcal{L} = -\frac{1}{B} \sum_{i=1}^{B} \log \frac{\sum_{j \in \mathcal{P}(i)} \exp(\text{sim}(f_i, f_j))}{\sum_{k=1}^{B} \mathbb{I}_{[k \ne i]} \exp(\text{sim}(f_i, f_k))},
\end{equation}

where \( f_i \) is log-prob gradient with respect to the final layer of sample \( x_i \) and \( \text{sim}(\cdot,\cdot) \) denotes cosine similarity. For a sample $i$, \( \mathcal{P}(i) \) is the set of unsimilar samples for \( i \). 

\section{Experiment}

\subsection{Experiment Setup}
\textbf{Benchmark}. To comprehensively evaluate the multimodal reasoning, perception, and hallucination resistance of large models, we adopt a diverse set of benchmarks covering spatial, temporal, and mathematical capabilities. MMVU~\cite{zhao2025mmvu} evaluates expert-level multidisciplinary video understanding. VideoMMMU~\cite{hu2025video} assesses knowledge-based question-answering over time. VideoHallucer~\cite{wang2024videohallucer} and POPE~\cite{li2023evaluating} evaluate resistance to hallucinations in video and image understanding, respectively. MathVista~\cite{lu2023mathvista} presents problems requiring logical, algebraic, and scientific inference, including quantitative reasoning. MMBench~\cite{liu2024mmbench} comprehensively assesses perception and cognitive abilities. This suite jointly evaluates models' visual, temporal, analytical, and hallucination-resistant capabilities across diverse scenarios.

\textbf{Baseline}. To evaluate our method, we compare it against a range of representative baselines. GPT-4o~\cite{hurst2024gpt} and Gemini-1.5-Pro~\cite{team2024gemini} serve as strong closed-source performance references. Open-source video models, including LLaVA-OneVision-7B~\cite{li2024llava} and VILA-1.5-8B, are relevant for temporal visual understanding. General-purpose vision-language foundations such as MiniCPM-V2.6-8B~\cite{yao2024minicpm}, InternVL2.5-8B~\cite{chen2024expanding}, and LLaMA3.2-11B~\cite{grattafiori2024llama} are used as base models. Reasoning-focused baselines like R1-OneVision-7B~\cite{yang2025r1}, R1-VL~\cite{zhang2025r1}, Vision-R1-7B~\cite{huang2025vision}, and Video-R1-7B~\cite{feng2025video} incorporate structured prompting or alignment for improved multi-step reasoning. Finally, we include our models: the base Qwen-VL-2.5-7B~\cite{bai2025qwen2}, its SFT and GRPO variants, and our enhanced version (Ours), to analyze the impact of visual grounding and contrastive regularization.

\subsection{Main Result}
\newcommand{\tablefontsize}{\small}

\begin{table*}[ht]
\centering
\small
\renewcommand{\arraystretch}{1.1}
\setlength{\tabcolsep}{5pt}
\begin{tabular}{clcccccc}
\toprule
\textbf{Category} & \textbf{Model} & \textbf{MMVU} & \textbf{VideoMMMU} & \textbf{VideoHallucer} & \textbf{MathVista} & \textbf{POPE} & \textbf{MMBench(en)} \\
\midrule

\multirow{2}{*}{Closed-source} 
  & GPT-4o                        & 75.4 & 61.2 & 53.3 & 61.3 & 86.9 & 83.4 \\
  & Gemini-1.5-Pro               & - & 60.6 & 37.8 & 63.9 & - & 73.9 \\
\midrule

\multirow{2}{*}{\makecell{Open-source \\ Video Models}} 
  & LLaVA-OneVision-7B          & 49.2 & 31.2 & 44.6 & 62.6 & 86.4 & 81.7 \\
  & VILA-1.5-8B                 & 31.5 & 20.8 & 14.8 & 36.7 & 70.6 & 57.6 \\
\midrule

\multirow{3}{*}{\makecell{Open-source \\ Base Models}} 
  & MiniCPM-V2.6-8B             & 52.4 & 49.8 & 48.4 & 60.6 & 84.4 & 81.5 \\
  & InternVL2.5-8B              & 54.9 & 44.2 & 50.5 & 64.4 & 85.9 & 84.6 \\
  & LLaMA3.2-11B                & - & - & - & 51.5 & 86.2 & 65.8 \\
\midrule

\multirow{4}{*}{\makecell{Open-source \\ Reasoning Models}} 
  & R1-OneVision-7B             & 55.2 & 44.1 & 42.4 & 64.1 & 83.1 & 82.3 \\
  & R1-VL-7B                    & 59.7 & 42.9 & 43.6 & 63.5 & 85.7 & 86.2 \\
  & Vision-R1-7B                & 57.6 & 39.7 & 44.2 & \textbf{73.5} & 86.4 & 83.8 \\
  & Video-R1-7B                 & 62.5 & 48.1 & 45.1 & 71.0 & 85.5 & 86.5 \\
\midrule

\multirow{4}{*}{\makecell{Our \\ Models}} 
  & Qwen-VL-2.5-7B              & 57.6 & 43.9 & 46.5 & 63.7 & 84.4 & 86.3 \\
  & Qwen-VL-2.5-7B-SFT          & 62.7 & 46.0 & 43.5 & 54.7 & 82.2 & 83.9 \\
  & Qwen-VL-2.5-7B-GRPO         & 62.1 & 47.3 & 42.3 & 69.3 & 83.6 & 86.8 \\
  & Qwen-VL-2.5-7B-Ours         & \textbf{65.6} & \textbf{50.0} & \textbf{50.8} & 69.4 & \textbf{88.7} & \textbf{88.6} \\
\bottomrule
\end{tabular}
\caption{Benchmark Performance Comparison of Closed-source, Open-source, and Proposed Models across Video, Reasoning, Math, and Hallucination-related Tasks. Bolded scores indicate the best performance among all open-source and our models.}\label{tab:model_comparison}
\end{table*}

As shown in Table~\ref{tab:model_comparison}, our model achieves state-of-the-art performance among all open-source models across image, video, and hallucination benchmarks, particularly excelling in video reasoning tasks. On MMVU—a knowledge-intensive video benchmark—our model attains 65.6\% accuracy, surpassing all open-source models and ranking second only to GPT-4o, highlighting its strong video inference capabilities. While performance on MathVista (focused on abstract mathematical reasoning) is moderate, our model delivers leading results on image-heavy benchmarks such as MMBench (88.6\%), demonstrating broad effectiveness in visual understanding.

In addition, standard GRPO training reduces performance on the hallucination benchmark, demonstrating that RL can introduce hallucinations. Our model achieves top results on VideoHallucer (50.8\%), indicating improved faithfulness and reduced hallucination risks compared to prior methods. Notably, supervised fine-tuning brings limited improvements, whereas RL yields consistent gains across all reasoning benchmarks, confirming its importance in enhancing both reasoning ability and response reliability.

To complement the main results, we evaluate whether the Caption Reward improves visual grounding, whether selecting moderate samples enhances exploration diversity in RL, and whether InfoNCE loss mitigates interference between conflicting samples.

\subsection{Effect of Caption Reward on Hallucination Reduction}
To evaluate the effectiveness of Caption Reward, we conduct an ablation study comparing models trained with and without this component across standard multimodal reasoning benchmarks. In particular, we aim to quantify its impact on hallucination reduction and visual grounding.

\begin{table}[ht]
\centering
\setlength{\tabcolsep}{2pt} 
\small                 
\begin{tabular}{lccc}
\toprule
\textbf{Method} & \textbf{MMVU} & \textbf{VideoHallucer} & \textbf{POPE} \\
\midrule
w/o Caption & 62.6 & 47.6 & 85.2 \\
w/o Caption Reward & 63.3 & 48.6 & 86.6 \\
w/ (Caption + Caption Reward) & \textbf{65.6} & \textbf{50.8} & \textbf{88.7} \\
\bottomrule
\end{tabular}
\caption{Impact of Caption and Caption Reward on Qwen-VL-2.5-7B Performance over MMVU, VideoHallucer and POPE.}
\label{tab:caption_ablation}
\end{table}

The results, as shown in Table \ref{tab:caption_ablation}, indicate that the model achieves the best performance when adopting our proposed output stages and incorporating the Caption Reward. Conversely, performance decreases when the Caption Reward is not applied, highlighting its effectiveness. Additionally, when the model does not follow our proposed output stages, performance further declines, underscoring the importance of the output stages we introduced.
\begin{figure}[ht]
    \centering
    \includegraphics[width=0.9\linewidth]{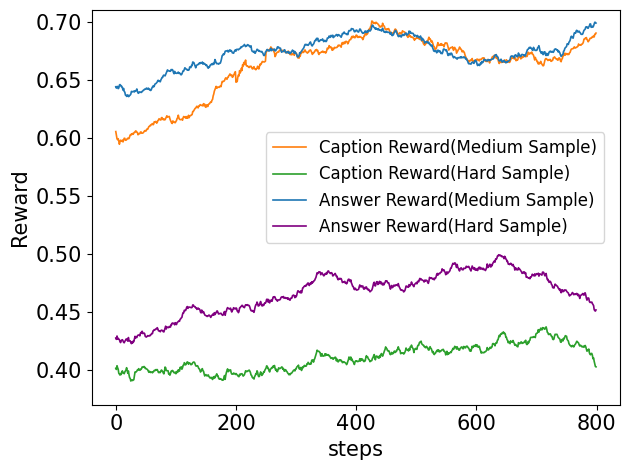}
    \caption{
Caption and Answer Reward curves during RL training on medium and hard samples.
}
    \label{fig:reward}
\end{figure}

To better understand how reward signals evolve during RL training, we visualize both Caption and Answer Rewards on medium and hard samples in Figure~\ref{fig:reward}. Caption Reward starts low, indicating weak visual grounding, but rises steadily on medium samples and gradually aligns with Answer Reward—suggesting that learning better captions directly supports accurate answering. In contrast, Caption Reward on hard samples improves slowly and remains low, limiting its effectiveness as a learning signal.

These trends highlight the central role of Caption Reward: it not only reflects visual grounding quality but also acts as a guiding signal that shapes the answer distribution. Its improvement precedes and facilitates gains in Answer Reward, especially on learnable (medium) samples, underscoring its value in promoting stable and effective multimodal learning.

\subsection{RL Exploration Diversity of Different Sample Types}

\begin{table}[ht]
\centering
\small
\begin{tabular}{lccc}
\toprule
\textbf{Sample Type} & \textbf{MMVU} & \textbf{VideoHallucer} & \textbf{POPE} \\
\midrule
Full Sample & 64.3 & 49.4 & 87.5 \\
Easy Sample & 62.2 & 47.7 & 86.6 \\
Hard Sample  & 63.5 & 48.4 & 88.2  \\
Medium Sample & \textbf{65.6} & \textbf{50.8} & \textbf{88.7} \\
\bottomrule
\end{tabular}
\caption{Impact of Sample Type on Qwen-VL-2.5-7B Performance over MMVU, VideoHallucer and POPE.}
\label{tab:type_ablation_full}
\end{table}
Table \ref{tab:type_ablation_full} indicates that the model trained with medium samples performs the best on all benchmarks, achieving results that are comparable to, or even superior to, those achieved by models trained on the full dataset. In contrast, training with easy samples leads to moderate performance due to the model's tendency to overfit, as it has a strong grasp of these samples. On the other hand, training with hard samples results in poor performance because the model struggles to comprehend these samples, preventing meaningful knowledge acquisition. For medium samples, however, the model is at a critical threshold where it can neither fully grasp nor completely fail to understand them. Through further reinforcement learning, the model successfully learns new knowledge from this subset, leading to superior performance.

\begin{figure}[h]
    \centering
    \includegraphics[width=0.9\linewidth]{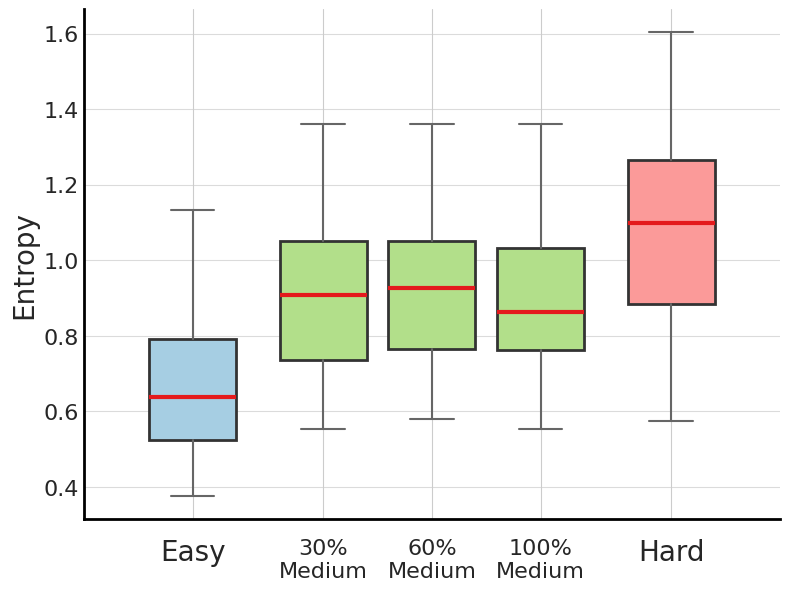}
    \caption{Policy entropy distribution across models trained on easy, medium, and hard samples.}
    \label{fig:xiangxing}
\end{figure}

To better understand how training samples influence model behavior, we analyze policy entropy as a proxy for action diversity. We randomly select 100 ChartQA~\cite{masry2022chartqa} test examples and compute the average token-level entropy for each. As shown in Figure~\ref{fig:xiangxing}, we compare models trained on easy, hard, and medium samples. For the medium category, we further sort samples by reward variance and construct three subsets containing the top 30\%, 60\%, and 100\% high-variance samples to examine the impact of targeted sample selection.

As illustrated in Figure~\ref{fig:xiangxing}, models trained solely on easy samples produce low-entropy outputs, indicating overconfident and potentially overfitted policies. Conversely, models trained on hard samples exhibit high entropy with large variability, suggesting unstable or unfocused decision-making due to the complexity of training inputs. Notably, the medium-sample models yield moderate entropy with a tighter distribution, indicating a balance between diversity and reliability. Moreover, models trained on high-variance subsets (30\% and 60\%) of medium samples achieve comparable or more stable entropy patterns than using the full medium set, suggesting that carefully selecting uncertain yet learnable samples within the medium group can further enhance training efficiency and policy quality.

\subsection{Evaluating the Effectiveness of InfoNCE Loss in Reducing Sample Conflicts}

\begin{table}[t]
\centering
\small
\begin{tabular}{lccc}
\toprule
\textbf{Method} & \textbf{MMVU} & \textbf{VideoHallucer} & \textbf{POPE} \\
\midrule
w/o InfoNCELoss & 63.8 & 48.3 & 86.8 \\
w/ InfoNCELoss & \textbf{65.6} & \textbf{50.8} & \textbf{88.7} \\
\bottomrule
\end{tabular}
\caption{Impact of InfoNCELoss on Qwen-VL-2.5-7B Performance over MMVU, VideoHallucer and POPE.}
\label{tab:infonce_ablation}
\end{table}

As shown in Table~\ref{tab:infonce_ablation}, incorporating InfoNCE Loss consistently improves performance across all benchmarks. These improvements highlight the effectiveness of InfoNCE in enhancing model robustness against hallucination.

\begin{figure}[ht]
    \centering
    \includegraphics[width=0.9\linewidth]{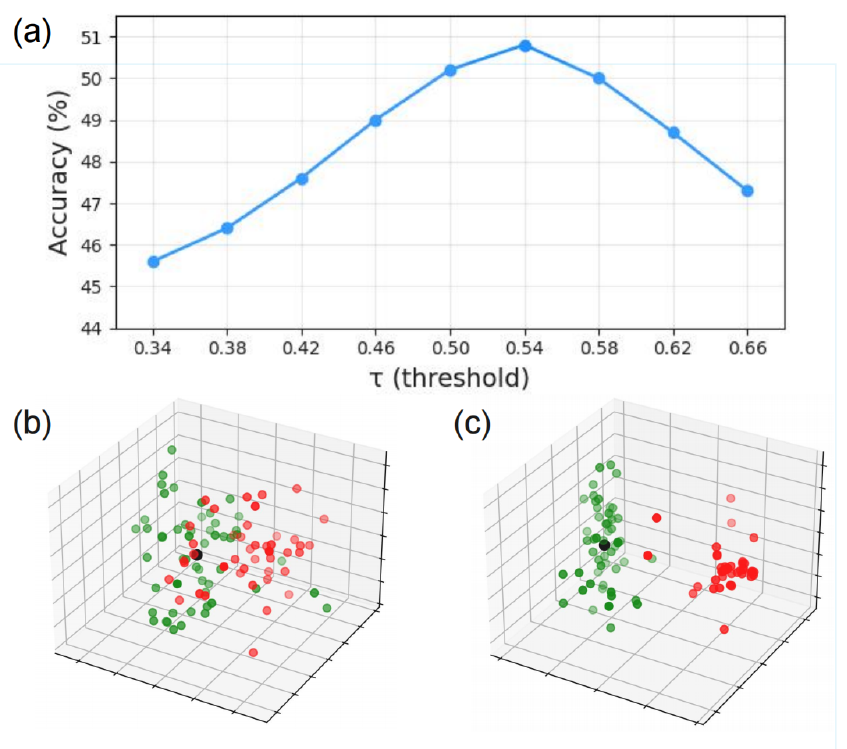}
    \caption{
(a) Accuracy under different NTK thresholds $\tau$. 
(b) and (c) Reshaping Effect on Final Layer Sample Representations Before and After InfoNCE Application.
}
    \label{fig:cluster}
\end{figure}

To better understand the mechanism behind the improvement, we first analyze the impact of the NTK similarity threshold $\tau$ used to guide the InfoNCE loss. As shown in Figure~\ref{fig:cluster}(a), model accuracy peaks when $\tau = 0.54$, indicating that this value provides a balanced criterion for distinguishing beneficial and conflicting training interactions. $\tau$ guides the InfoNCE Loss in learning to distinguish between samples that facilitate or hinder learning.

We then investigate how this NTK-guided contrastive loss reshapes the model’s internal representations. Specifically, we randomly select 129 samples and designate one as the anchor. Based on their NTK similarity to the anchor and the chosen threshold $\tau$, we construct positive and negative pairs for the InfoNCE Loss. Figure~\ref{fig:cluster}(b) and (c) show the final-layer representations before and after training. Prior to training, the samples are dispersed without clear structure. After training, cooperative samples (green), treated as positives—form a compact neighborhood around the anchor, while conflicting samples (red)—treated as negatives—are pushed apart. This structured separation emerges from contrastive optimization rather than from the initial threshold itself, and reflects improved functional organization and generalization.

\section{Conclusion}
This work presents a systematic study of hallucination in MLLMs trained with RL. We identify three primary causes—visual representation errors, limited exploration diversity, and sample-level conflicts—that undermine the reliability of model reasoning. To tackle these challenges, we introduce three components: a Caption Reward to enhance visual grounding, a reward-variance-based sampling strategy to encourage diverse exploration, and contrastive regularization to reduce interference between training samples. Our approach enhances visual faithfulness, improves robustness, and stabilizes learning dynamics of MLLMs without compromising generalization. Extensive evaluations on image and video benchmarks demonstrate that the proposed framework consistently reduces hallucination and strengthens multimodal reasoning under RL training.


\newpage
\section{Acknowledgments}
This work is supported by National Key R\&D Program of China under Grant No. 2024YFB3908400. This work is also supported by the Key R\&D Program of Ningbo under Grant No.2024Z115.

\bibliography{aaai2026}

@article{zhu2025surprising,
  title={The Surprising Effectiveness of Negative Reinforcement in LLM Reasoning},
  author={Zhu, Xinyu and Xia, Mengzhou and Wei, Zhepei and Chen, Wei-Lin and Chen, Danqi and Meng, Yu},
  journal={arXiv preprint arXiv:2506.01347},
  year={2025}
}

@article{liu2025more,
  title={More Thinking, Less Seeing? Assessing Amplified Hallucination in Multimodal Reasoning Models},
  author={Liu, Chengzhi and Xu, Zhongxing and Wei, Qingyue and Wu, Juncheng and Zou, James and Wang, Xin Eric and Zhou, Yuyin and Liu, Sheng},
  journal={arXiv preprint arXiv:2505.21523},
  year={2025}
}

@article{lee2025retrieval,
  title={Retrieval Visual Contrastive Decoding to Mitigate Object Hallucinations in Large Vision-Language Models},
  author={Lee, Jihoon and Song, Min},
  journal={arXiv preprint arXiv:2505.20569},
  year={2025}
}

@article{ghosh2024visual,
  title={Visual description grounding reduces hallucinations and boosts reasoning in lvlms},
  author={Ghosh, Sreyan and Evuru, Chandra Kiran Reddy and Kumar, Sonal and Tyagi, Utkarsh and Nieto, Oriol and Jin, Zeyu and Manocha, Dinesh},
  journal={arXiv preprint arXiv:2405.15683},
  year={2024}
}

@inproceedings{leng2024mitigating,
  title={Mitigating object hallucinations in large vision-language models through visual contrastive decoding},
  author={Leng, Sicong and Zhang, Hang and Chen, Guanzheng and Li, Xin and Lu, Shijian and Miao, Chunyan and Bing, Lidong},
  booktitle={Proceedings of the IEEE/CVF Conference on Computer Vision and Pattern Recognition},
  pages={13872--13882},
  year={2024}
}

@article{hu2025video,
  title={Video-MMMU: Evaluating Knowledge Acquisition from Multi-Discipline Professional Videos},
  author={Hu, Kairui and Wu, Penghao and Pu, Fanyi and Xiao, Wang and Zhang, Yuanhan and Yue, Xiang and Li, Bo and Liu, Ziwei},
  journal={arXiv preprint arXiv:2501.13826},
  year={2025}
}

@article{lu2023mathvista,
  title={Mathvista: Evaluating mathematical reasoning of foundation models in visual contexts},
  author={Lu, Pan and Bansal, Hritik and Xia, Tony and Liu, Jiacheng and Li, Chunyuan and Hajishirzi, Hannaneh and Cheng, Hao and Chang, Kai-Wei and Galley, Michel and Gao, Jianfeng},
  journal={arXiv preprint arXiv:2310.02255},
  year={2023}
}

@article{masry2022chartqa,
  title={Chartqa: A benchmark for question answering about charts with visual and logical reasoning},
  author={Masry, Ahmed and Long, Do Xuan and Tan, Jia Qing and Joty, Shafiq and Hoque, Enamul},
  journal={arXiv preprint arXiv:2203.10244},
  year={2022}
}

@article{hurst2024gpt,
  title={Gpt-4o system card},
  author={Hurst, Aaron and Lerer, Adam and Goucher, Adam P and Perelman, Adam and Ramesh, Aditya and Clark, Aidan and Ostrow, AJ and Welihinda, Akila and Hayes, Alan and Radford, Alec and others},
  journal={arXiv preprint arXiv:2410.21276},
  year={2024}
}

@article{team2024gemini,
  title={Gemini 1.5: Unlocking multimodal understanding across millions of tokens of context},
  author={Team, Gemini and Georgiev, Petko and Lei, Ving Ian and Burnell, Ryan and Bai, Libin and Gulati, Anmol and Tanzer, Garrett and Vincent, Damien and Pan, Zhufeng and Wang, Shibo and others},
  journal={arXiv preprint arXiv:2403.05530},
  year={2024}
}

@article{li2024llava,
  title={Llava-onevision: Easy visual task transfer},
  author={Li, Bo and Zhang, Yuanhan and Guo, Dong and Zhang, Renrui and Li, Feng and Zhang, Hao and Zhang, Kaichen and Zhang, Peiyuan and Li, Yanwei and Liu, Ziwei and others},
  journal={arXiv preprint arXiv:2408.03326},
  year={2024}
}

@article{yao2024minicpm,
  title={Minicpm-v: A gpt-4v level mllm on your phone},
  author={Yao, Yuan and Yu, Tianyu and Zhang, Ao and Wang, Chongyi and Cui, Junbo and Zhu, Hongji and Cai, Tianchi and Li, Haoyu and Zhao, Weilin and He, Zhihui and others},
  journal={arXiv preprint arXiv:2408.01800},
  year={2024}
}

@article{chen2024expanding,
  title={Expanding performance boundaries of open-source multimodal models with model, data, and test-time scaling},
  author={Chen, Zhe and Wang, Weiyun and Cao, Yue and Liu, Yangzhou and Gao, Zhangwei and Cui, Erfei and Zhu, Jinguo and Ye, Shenglong and Tian, Hao and Liu, Zhaoyang and others},
  journal={arXiv preprint arXiv:2412.05271},
  year={2024}
}

@article{grattafiori2024llama,
  title={The llama 3 herd of models},
  author={Grattafiori, Aaron and Dubey, Abhimanyu and Jauhri, Abhinav and Pandey, Abhinav and Kadian, Abhishek and Al-Dahle, Ahmad and Letman, Aiesha and Mathur, Akhil and Schelten, Alan and Vaughan, Alex and others},
  journal={arXiv preprint arXiv:2407.21783},
  year={2024}
}

@article{yang2025r1,
  title={R1-onevision: Advancing generalized multimodal reasoning through cross-modal formalization},
  author={Yang, Yi and He, Xiaoxuan and Pan, Hongkun and Jiang, Xiyan and Deng, Yan and Yang, Xingtao and Lu, Haoyu and Yin, Dacheng and Rao, Fengyun and Zhu, Minfeng and others},
  journal={arXiv preprint arXiv:2503.10615},
  year={2025}
}

@article{zhang2025r1,
  title={R1-vl: Learning to reason with multimodal large language models via step-wise group relative policy optimization},
  author={Zhang, Jingyi and Huang, Jiaxing and Yao, Huanjin and Liu, Shunyu and Zhang, Xikun and Lu, Shijian and Tao, Dacheng},
  journal={arXiv preprint arXiv:2503.12937},
  year={2025}
}

@article{huang2025vision,
  title={Vision-r1: Incentivizing reasoning capability in multimodal large language models},
  author={Huang, Wenxuan and Jia, Bohan and Zhai, Zijie and Cao, Shaosheng and Ye, Zheyu and Zhao, Fei and Xu, Zhe and Hu, Yao and Lin, Shaohui},
  journal={arXiv preprint arXiv:2503.06749},
  year={2025}
}

@article{bai2025qwen2,
  title={Qwen2. 5-vl technical report},
  author={Bai, Shuai and Chen, Keqin and Liu, Xuejing and Wang, Jialin and Ge, Wenbin and Song, Sibo and Dang, Kai and Wang, Peng and Wang, Shijie and Tang, Jun and others},
  journal={arXiv preprint arXiv:2502.13923},
  year={2025}
}

@article{shao2024deepseekmath,
  title={Deepseekmath: Pushing the limits of mathematical reasoning in open language models},
  author={Shao, Zhihong and Wang, Peiyi and Zhu, Qihao and Xu, Runxin and Song, Junxiao and Bi, Xiao and Zhang, Haowei and Zhang, Mingchuan and Li, YK and Wu, Y and others},
  journal={arXiv preprint arXiv:2402.03300},
  year={2024}
}

@article{wang2024videohallucer,
  title={Videohallucer: Evaluating intrinsic and extrinsic hallucinations in large video-language models},
  author={Wang, Yuxuan and Wang, Yueqian and Zhao, Dongyan and Xie, Cihang and Zheng, Zilong},
  journal={arXiv preprint arXiv:2406.16338},
  year={2024}
}

@inproceedings{zhao2025mmvu,
  title={Mmvu: Measuring expert-level multi-discipline video understanding},
  author={Zhao, Yilun and Zhang, Haowei and Xie, Lujing and Hu, Tongyan and Gan, Guo and Long, Yitao and Hu, Zhiyuan and Chen, Weiyuan and Li, Chuhan and Xu, Zhijian and others},
  booktitle={Proceedings of the Computer Vision and Pattern Recognition Conference},
  pages={8475--8489},
  year={2025}
}

@inproceedings{liu2024mmbench,
  title={Mmbench: Is your multi-modal model an all-around player?},
  author={Liu, Yuan and Duan, Haodong and Zhang, Yuanhan and Li, Bo and Zhang, Songyang and Zhao, Wangbo and Yuan, Yike and Wang, Jiaqi and He, Conghui and Liu, Ziwei and others},
  booktitle={European conference on computer vision},
  pages={216--233},
  year={2024},
  organization={Springer}
}

@article{Kaelbling_Littman_Moore_2018,  
 title={Reinforcement Learning: A Survey}, 
 url={http://dx.doi.org/10.1613/jair.301}, 
 DOI={10.1613/jair.301}, 
 journal={Journal of Artificial Intelligence Research}, 
 author={Kaelbling, L. P. and Littman, M. L. and Moore, A. W.}, 
 year={2018}, 
 month={Jul}, 
 pages={237–285}, 
 language={en-US} 
 }

@article{jacot2018neural,
  title={Neural tangent kernel: Convergence and generalization in neural networks},
  author={Jacot, Arthur and Gabriel, Franck and Hongler, Cl{\'e}ment},
  journal={Advances in neural information processing systems},
  volume={31},
  year={2018}
}

@article{peng2025lmm,
  title={Lmm-r1: Empowering 3b lmms with strong reasoning abilities through two-stage rule-based rl},
  author={Peng, Yingzhe and Zhang, Gongrui and Zhang, Miaosen and You, Zhiyuan and Liu, Jie and Zhu, Qipeng and Yang, Kai and Xu, Xingzhong and Geng, Xin and Yang, Xu},
  journal={arXiv preprint arXiv:2503.07536},
  year={2025}
}

@article{feng2025video,
  title={Video-r1: Reinforcing video reasoning in mllms},
  author={Feng, Kaituo and Gong, Kaixiong and Li, Bohao and Guo, Zonghao and Wang, Yibing and Peng, Tianshuo and Wu, Junfei and Zhang, Xiaoying and Wang, Benyou and Yue, Xiangyu},
  journal={arXiv preprint arXiv:2503.21776},
  year={2025}
}

@article{li2025q,
  title={Q-insight: Understanding image quality via visual reinforcement learning},
  author={Li, Weiqi and Zhang, Xuanyu and Zhao, Shijie and Zhang, Yabin and Li, Junlin and Zhang, Li and Zhang, Jian},
  journal={arXiv preprint arXiv:2503.22679},
  year={2025}
}

@article{dona2025bettercheck,
  title={BetterCheck: Towards Safeguarding VLMs for Automotive Perception Systems},
  author={Dona, Malsha Ashani Mahawatta and Cabrero-Daniel, Beatriz and Yu, Yinan and Berger, Christian},
  journal={arXiv preprint arXiv:2507.17722},
  year={2025}
}

@article{yao2025reasoning,
  title={Are Reasoning Models More Prone to Hallucination?},
  author={Yao, Zijun and Liu, Yantao and Chen, Yanxu and Chen, Jianhui and Fang, Junfeng and Hou, Lei and Li, Juanzi and Chua, Tat-Seng},
  journal={arXiv preprint arXiv:2505.23646},
  year={2025}
}

@article{pan2025medvlm,
  title={Medvlm-r1: Incentivizing medical reasoning capability of vision-language models (vlms) via reinforcement learning},
  author={Pan, Jiazhen and Liu, Che and Wu, Junde and Liu, Fenglin and Zhu, Jiayuan and Li, Hongwei Bran and Chen, Chen and Ouyang, Cheng and Rueckert, Daniel},
  journal={arXiv preprint arXiv:2502.19634},
  year={2025}
}

@article{zhou2025r1,
  title={R1-Zero's" Aha Moment" in Visual Reasoning on a 2B Non-SFT Model},
  author={Zhou, Hengguang and Li, Xirui and Wang, Ruochen and Cheng, Minhao and Zhou, Tianyi and Hsieh, Cho-Jui},
  journal={arXiv preprint arXiv:2503.05132},
  year={2025}
}

@article{meng2025mm,
  title={Mm-eureka: Exploring visual aha moment with rule-based large-scale reinforcement learning},
  author={Meng, Fanqing and Du, Lingxiao and Liu, Zongkai and Zhou, Zhixiang and Lu, Quanfeng and Fu, Daocheng and Shi, Botian and Wang, Wenhai and He, Junjun and Zhang, Kaipeng and others},
  journal={CoRR},
  year={2025}
}

@article{liu2025othink,
  title={OThink-MR1: Stimulating multimodal generalized reasoning capabilities via dynamic reinforcement learning},
  author={Liu, Zhiyuan and Zhang, Yuting and Liu, Feng and Zhang, Changwang and Sun, Ying and Wang, Jun},
  journal={arXiv preprint arXiv:2503.16081},
  year={2025}
}

@article{liu2025seg,
  title={Seg-zero: Reasoning-chain guided segmentation via cognitive reinforcement},
  author={Liu, Yuqi and Peng, Bohao and Zhong, Zhisheng and Yue, Zihao and Lu, Fanbin and Yu, Bei and Jia, Jiaya},
  journal={arXiv preprint arXiv:2503.06520},
  year={2025}
}

@article{pan2025metaspatial,
  title={Metaspatial: Reinforcing 3d spatial reasoning in vlms for the metaverse},
  author={Pan, Zhenyu and Liu, Han},
  journal={arXiv preprint arXiv:2503.18470},
  year={2025}
}

@inproceedings{kang2025gflowvlm,
  title={GFlowVLM: Enhancing Multi-step Reasoning in Vision-Language Models with Generative Flow Networks},
  author={Kang, Haoqiang and Sachdeva, Enna and Gupta, Piyush and Bae, Sangjae and Lee, Kwonjoon},
  booktitle={Proceedings of the Computer Vision and Pattern Recognition Conference},
  pages={3815--3825},
  year={2025}
}

@article{deng2025boosting,
  title={Boosting the generalization and reasoning of vision language models with curriculum reinforcement learning},
  author={Deng, Huilin and Zou, Ding and Ma, Rui and Luo, Hongchen and Cao, Yang and Kang, Yu},
  journal={arXiv preprint arXiv:2503.07065},
  year={2025}
}

@article{lu2025ui,
  title={UI-R1: Enhancing Efficient Action Prediction of GUI Agents by Reinforcement Learning},
  author={Lu, Zhengxi and Chai, Yuxiang and Guo, Yaxuan and Yin, Xi and Liu, Liang and Wang, Hao and Xiao, Han and Ren, Shuai and Xiong, Guanjing and Li, Hongsheng},
  journal={arXiv preprint arXiv:2503.21620},
  year={2025}
}

@article{christiano2017deep,
  title={Deep reinforcement learning from human preferences},
  author={Christiano, Paul F and Leike, Jan and Brown, Tom and Martic, Miljan and Legg, Shane and Amodei, Dario},
  journal={Advances in neural information processing systems},
  volume={30},
  year={2017}
}

@article{li2023evaluating,
  title={Evaluating object hallucination in large vision-language models},
  author={Li, Yifan and Du, Yifan and Zhou, Kun and Wang, Jinpeng and Zhao, Wayne Xin and Wen, Ji-Rong},
  journal={arXiv preprint arXiv:2305.10355},
  year={2023}
}


\end{document}